\documentclass[letterpaper, 10 pt, conference]{ieeeconf}  
\pdfoutput=1
\usepackage{graphicx}
\usepackage[abs]{overpic}
\usepackage{amsmath}
\usepackage{amsfonts}
\usepackage{amssymb}
\usepackage{multirow}
\usepackage{xcolor}
\usepackage{grffile}
\usepackage{hyperref}

\IEEEoverridecommandlockouts                              

\title{\LARGE \bf
Towards Target-Driven Visual Navigation in Indoor Scenes via Generative Imitation Learning}

\author{Qiaoyun~Wu,
Xiaoxi~Gong,
Kai~Xu,
Dinesh~Manocha,
Jingxuan~Dong,
Jun~Wang\thanks{Corresponding author: Jun Wang (wjun@nuaa.edu.cn)}
\thanks{Q.~Wu, X.~Gong, J.~Wang and J.~Dong are with the College of Mechanical and Electrical Engineering,
Nanjing University of Aeronautics and Astronautics, China.}
\thanks{K.~Xu is with the School of Computer Science, National University of Defense Technology, China.}
\thanks{D. Manocha is with the Department of Computer Science, the University of Maryland, College Park.}
\thanks{This article has supplementary downloadable material available at https://ieeexplore.ieee.org/document/9250507, provided by the authors.}
\thanks{Digital Object Identifier (DOI): 10.1109/LRA.2020.3036597}
}

\begin{document}

\maketitle
\thispagestyle{empty}
\pagestyle{empty}

\begin{abstract}
We present a target-driven navigation system to improve mapless visual navigation in indoor scenes.
Our method takes a multi-view observation of a robot and a target as inputs at each time step to provide a sequence of actions that move the robot to the target
without relying on odometry or GPS at runtime.
The system is learned by optimizing a combinational objective encompassing three key designs.
First, we propose that an agent conceives the next observation before making an action decision. This is achieved by learning a variational generative module from expert demonstrations.
We then propose predicting static collision in advance, as an auxiliary task
to improve safety during navigation.
Moreover, to alleviate the training data imbalance problem of termination action prediction, we also introduce a target checking module to differentiate from augmenting navigation policy with a termination action.
The three proposed designs all contribute to the improved training data efficiency, static collision avoidance, and navigation generalization performance, resulting in a novel target-driven mapless navigation system.
Through experiments on a TurtleBot, we provide evidence that our model can be integrated into a robotic system and navigate in the real world.
\emph{Videos and models can be found in the supplementary material\footnote{https://github.com/wqynew/Enhanced-NeoNav}.}
\end{abstract}

\section{Introduction}
The last decade has seen significant achievements in the field of autonomous navigation technologies,
starting with motion planning given a geometric model of the environment~\cite{lavalle2006planning},
then progressively integrating automation technologies into robots to
assist navigating in explored scenes~\cite{dame2011new,zhu2017,chen2019learning}.
However, the autonomous mobility of robots is still limited in an unexplored scene with a new navigation task, which greatly limits the mobile robot application in many tasks, including household service and restaurant delivery.

Traditionally, robotic navigation methods consist of two parts.
First, a geometric map is built using mapping techniques, such as Simultaneous Localization and Mapping (SLAM)~\cite{cadena2016past}. Next, a collision-free path in workspace or configuration space is sought with respect to the map using path planning algorithms, such as Probabilistic Roadmaps (PRM)~\cite{kavraki1996probabilistic} and Rapidly Exploring Random Trees (RRT)~\cite{lavalle2001rapidly}. However, these methods are highly sensitive to robot odometry and noise in sensor data. Representations constructed by SLAM systems are prone to error when the environment changes over time.
Motion planning approaches often assume perfect localization and rely on high-quality geometric maps of the environment.
More importantly, these methods ignore the rich information from on-board visual sensors of robots limiting the use of these methods to target-driven navigation (i.e., autonomously navigating to a semantic object).

Given the above limitations, deep learning-based mapless navigation approaches
have gained considerable attention recently.
These methods do not rely on prior knowledge of surroundings.
They predict navigation actions directly from visual observations of robots based on end-to-end learning, including Imitation Learning (IL)~\cite{Pomerleau1993,lecun2005,zhu2018scores,watkins2019learning} and Reinforcement Learning (RL)~\cite{fan2018crowdmove,gordon2019,fan2019learning,Sathyamoorthy2020}. IL based navigation requires the optimal demonstration from experts and has the advantage of fast learning of useful information~\cite{watkins2019learning}.
RL based navigation does not specifically require supervision by an expert, as it
searches for an optimal policy that finally leads to the highest reward.
However, it generally requires abundant training data to converge, suffers from sparse rewards in navigation episodes, and struggles to generalize to unseen scenes with new targets.

\begin{figure*}[thpb]
\begin{center}
\includegraphics[width=0.9\linewidth]{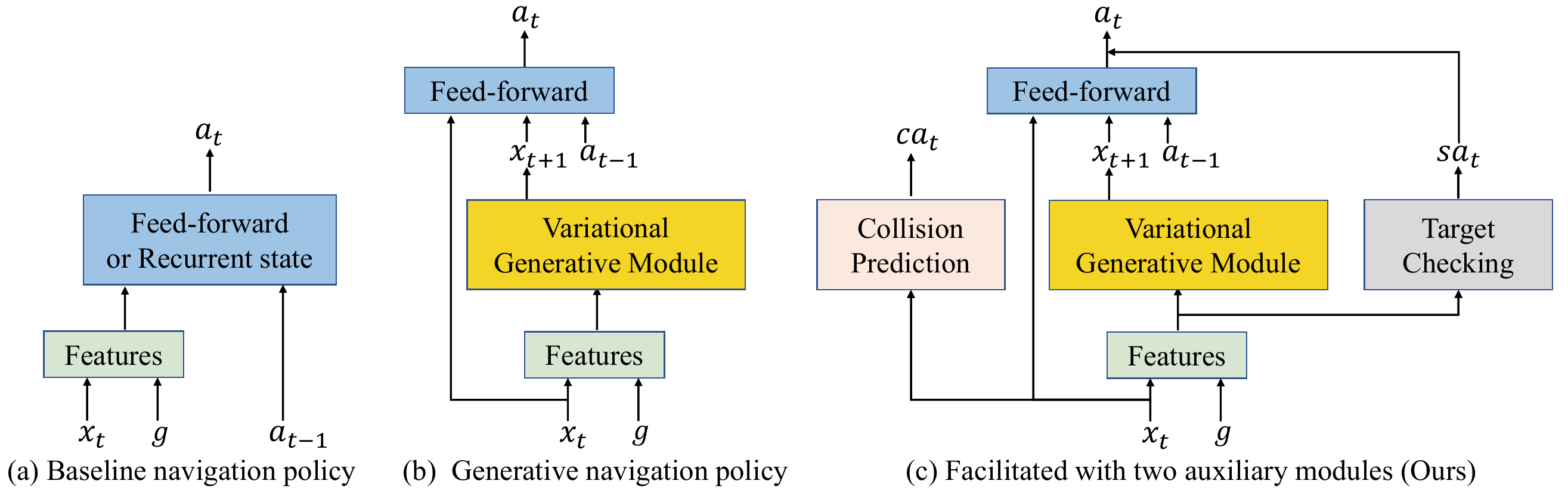}
\end{center}
\caption{Target-driven navigation takes as input the current and the target observations and outputs an action that would lead to the target. We compare the following navigation models: (a) Baseline navigation policy; (b) Generative navigation policy; (c) Our integrated navigation pipeline.
}
\label{fig:frame}\vspace{-10pt}
\end{figure*}

In this paper, we focus on exploring supervised methods (in particular, imitation learning) to bring better cross-scene and cross-target generalization to  target-driven visual navigation of robots.
Given the sequence of observations and actions from a demonstration task, our
navigation policy learns how to reach a target by imitating the expert demonstration step-by-step.
One critical challenge in learning the navigation policy is that, in general, there may be multiple possible ways of going from the current location to the target: that is, the distribution of trajectories between states is multi-modal~\cite{pathak2018zero}.
We address this issue with our novel variational generative module based on the idea of conceiving the next expected observation (NEO) before making an action decision.
To operationalize this, we first learn a generative model conditioned
on the multi-view observations at the current location as well as the target image, from which the NEO can be generated.
We predict the next action based on the difference between the generated NEO and the current (front-view) observation.
See Figure~\ref{fig:frame}(b) for a schematic illustration of the framework.
This framework has the effect of transferring the multi-modality of navigation action prediction to the generation of NEO, making the progress of action prediction a surjection instead of a multi-modal action distribution learning,
thus greatly enhancing data efficiency.
In addition, our NEO generation essentially models the forward dynamics of the agent-environment interaction, i.e. action-driven state transition.
This enables multiple ground truth actions to take effect in the generation of NEO,
improving the expressiveness of our generation module,
and thus facilitating the generalization performance.

In addition, we also incorporate recent insights relating network conditioning to navigation performance. Considering the learning efficiency and computation configuration,
we explore different network architectures and finally present the architecture with a good tradeoff.
To account for the static collision during robot navigation, we propose jointly optimizing the navigation policy with a premature collision prediction that evaluates the collision probability of all possible actions for every current position.
Furthermore, we also design a target checking module for deciding whether the robot has reached the target.
We will show that our method jointly trained with the module consistently outperforms the baseline, which augments the action space with a \emph{stop} action.

In summary, our contributions are as follows: (1) We present a navigation pipeline (see Figure~\ref{fig:frame}(c)) for navigating to novel targets in unexplored scenes using only the current visual observation and the target image, without relying on any location services at runtime. (2) We integrate a variational generative model into navigation policy learning, which strengthens the connection between robotic observation and navigation actions and helps alleviate the multi-modality in action decision making.
(3) We propose adding a premature collision prediction module downstream of the convolutional neural network of our original architecture to provide a strong learning signal that encourages learning of useful features for both navigation tasks and static collision avoidance. (4) We design a target checking module in response to an optimization on our original architecture, which has the policy model output a \emph{stop} action when a robot is close to a target position. Integrating with the module, our method demonstrates better navigation performance.

A preliminary version of the navigation model is presented in~\cite{wuneonav},
which proposes a generative model for visual navigation.
In this work, we have significantly extended the idea behind the model design by taking into account the multi-modality during navigation decision making, which is generally an important factor that affects the performance of navigation policy learning.
In addition, we investigate three techniques to improve robot navigation performance in the real world, including feature space dynamics, premature collision prediction and additional target checking.
We show that the proposed method significantly outperforms prior work~\cite{wuneonav},
boosting the success rate from $17.5\%$ to $28.7\%$ and reducing approximately $16.3\%$ of the collisions for a navigation task in the unseen scenes from the Active Vision Dataset~\cite{mousavian2019visual}.
Furthermore, we steer a wheeled robot, TurtleBot, around office scenes and show that the learned navigation policy can generalize to novel targets in unseen real-world environments.
\emph{Demonstration videos and the code can be found in the supplementary material.}

The remainder of this paper is organized as follows.
In Section~\ref{sec:related_work}, we review the relevant background literature.
Section~\ref{sec:pro} describes the target-driven visual navigation problem.
In Section~\ref{sec:policy}, we pose and solve the problem by integrating a variational generative model into navigation policy learning.
Section~\ref{sec:tech} presents three techniques to facilitate learning.
Section~\ref{sec:result} provides an exhaustive experimental validation of our designs.
We conclude in Section\ref{sec:con} with a summary and a discussion of future work.


\section{Related Work}
\label{sec:related_work}
The task of learning an agent (e.g. ground vehicle, UAV, or mobile robot) to physically navigate through an unknown environment has been approached either through reinforcement learning (RL) or imitation learning (IL). In this section, we provide a brief review of these related learning strategies for the sequential decision making problem of navigation.

\textbf{Reinforcement Learning.}
Reinforcement learning has achieved state-of-the-art results in different fields by directly maximising cumulative reward without counting on expert supervision.
Recently, a growing number of methods have been reported for RL-based navigation~\cite{wortsman2019,chiang2019learning,chen2020robot}.
For example, Zhu et al.~\cite{zhu2017} propose an architecture for target-driven visual navigation by combining a Siamese network with an A3C algorithm.
Ye et al.~\cite{ye2018active} focus on learning policies for robots to allow object searching
and reaching.
However, neither work considers the generalization to previously unseen environments.
Work in~\cite{mirowski2016} provides several additional RL learning strategies and associated architectures.
Wu et al.~\cite{wu2018building} focus on room navigation, in which an agent learns to understand a given semantic room concept and finally navigate to the target room.
The method shows strong result in some unseen environments of House3D.
Wortsman et al.~\cite{wortsman2019} propose a self-adaptive visual navigation (SAVN) method which learns to adapt to new environments on AI2-THOR without considering the generalization to novel targets.
Furthermore, many recent works have extended deep RL methods to real-world robotics applications by either collecting an exhaustive real-world dataset of simple maze-like environments under grid world assumptions~\cite{zhang2017}, or directly transferring a navigation model in simulation to real maze environments~\cite{devo2020towards}.
Anderson et al.~\cite{anderson2018evaluation} and Savva et al.~\cite{savva2019habitat} design RL-based agents for point navigation in realistic cluttered environments, which require an idealized GPS and the specific location of the goal at runtime.
We also evaluate our target-driven navigation model on real-world complex scenes, each
containing visually and structurally different observations, but without relying on any maps and localization devices.

\textbf{Inverse Reinforcement Learning.}
Inverse RL has recently been the most commonly used method~\cite{levine2016learning,gao2018reinforcement}.
The DAGGER model~\cite{ross2011} proposes continuously closing the trajectory distributions from the agent and the expert demonstration and has been widely used for many robotic control tasks.
To avoid directly interacting with the expert as in DAGGER,
Ho et al.~\cite{ho2016generative} design a generative adversarial model to fit distributions of states and actions defining expert behaviors.
These methods demonstrate higher sample efficiency and generalization than many classical RL methods.
Ziebart et al.~\cite{Ziebart2008} propose the maximum entropy inverse reinforcement learning (MaxEnt IRL), which is computationally efficient on a routing problem (mission
planning).
You et al.~\cite{You2019} learn the optimal driving strategy using inverse reinforcement learning based on the demonstrations from expert drivers, which demonstrates desired driving behaviors in some simulation environments.
Xia et al.~\cite{xia2016neural} focus on a specific navigation task and propose learning the underlying rewards from expert demonstrations under the framework of inverse RL.
The navigation target information is hard-coded in the neural networks, which does not support the cross-target generalization.
In visual navigation, Gupta et al.~\cite{gupta2017} present an end-to-end architecture based on DAGGER to jointly train mapping and planning for navigation in novel environments. One limitation of the work is the assumption of perfect odometry, which is not accessible in the real world. We propose a target-driven navigation system without relying on any topological maps or location measurements.

\textbf{Imitation Learning.}
Imitation learning (IL) aims to mimic human behavior by learning from demonstrations
~\cite{argall2009survey,lind2018deep,watkins2019learning}.
Richter et al.~\cite{richter2017safe} use a conventional feed forward neural network to predict collisions for robotic navigation based on images observed by the robot, which relies on an odometry and the preset goal location.
Codevilla el al.~\cite{Codevilla2018} propose a framework that learns sub-policies using a multi-headed network in the autonomous driving setting.
In~\cite{xu2018}, the authors propose a deep multi-task shared imitation learning framework, SMIL, that can learn to work on multiple robotics tasks with multiple sub-policies.
Mousavian et al.~\cite{mousavian2019visual} learn to predict the cost of an action, which is supervised by the shortest paths of navigation tasks.
Pfeiffer et al.~\cite{pfeiffer2018reinforced} leverage prior expert demonstrations for pre-training of laser-based navigation policy.
Pathak et al.~\cite{pathak2018zero} learn an inverse dynamics model based on the demonstrated trajectory way-points from the expert, which requires several intermediate sub-goals for a long-range navigation task.
Watkins et al.~\cite{watkins2019learning} train an agent to navigate to any position via direct behavioral cloning from pre-generated expert trajectories, given a panoramic view of the goal and the current visual input. However, an environment map should be given when generalizing to unseen environments.
In contrast to this work, we focus on both cross-target and cross-scene generalization for navigation and propose conceiving the next observation before acting and other techniques for optimization that make a more effective and generalizable navigation model.

\section{PROBLEM FORMULATION}
\label{sec:pro}
The goal of target-driven navigation is to learn a controller, which enables a robot to autonomously and safely navigate to a target in an unexplored scene, without providing any map, odometry, GPS or relative location of the target but only RGB or Depth input from on-board visual sensors.
The navigation target is described by an image, which is also specified as an input to our model.
Hence, for testing, a mobile robot with our model can navigate to new targets without re-training.

To achieve this, learning needs to be done through repeated interactions between an agent and an environment $\mathcal{E}$. At every time step $t$, the agent receives an observation $x_t$ from $\mathcal{E}$ and then performs an action $a_t$ within the available action space $\mathcal{A}$ based on its current policy $\pi (x_t, g)$, where $g$ is the target. Subsequently, the agent transfers to a new observation $x_{t+1}$ within the observation space $\mathcal{O}$ under the environment transition distribution $p(x_{t+1}|x_{t},a_{t})$. After repeating this process, the agent generates a trajectory $\{x_1, a_1,x_2,a_2,\cdots,x_T\}$, also named an episode.
An episode can end when the agent acts for a certain number of time steps or reaches the target.


\begin{figure}[thpb]
\begin{center}
\includegraphics[width=\linewidth]{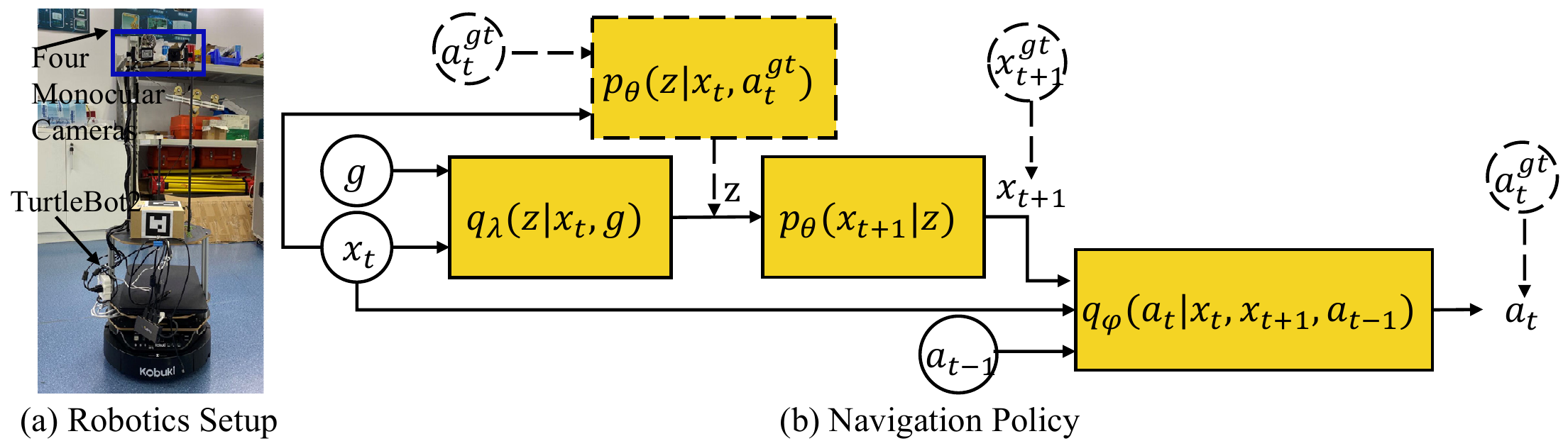}
\end{center}
   \caption{(a) The robotics system setup. (b) Our navigation policy mainly consists of the four components in  yellow squares. Symbols in solid circles denote the input and symbols in dotted circles represent the supervision from an expert to help update the parameters of the proposed policy.}
\label{fig:policy}
\end{figure}

We configure four RGB or Depth cameras to have a panoramic field of view $x_t$ at each time step. They are arranged at $0^\circ$, $90^\circ$, $180^\circ$ and $270^\circ$ horizontally, covering $90^\circ$ vertical fields of view (see Figure~\ref{fig:policy}(a)). The target $g$ is consistent with the observation $x_t$ in terms of image data modality.
We define a set of control commands: $\mathcal{A}=\{\emph{move~forward; move~back; move~left; move~right; rotate~ccw;} \\\emph{rotate~cw; stop}\}$. The \emph{rotate} \emph{ccw}$/$\emph{cw} action indicates turning the agent in place left$/$right $30^\circ$ and the \emph{move} action moves the agent a settled distance (e.g., $0.5m$). In our setting, an episode is terminated when the \emph{stop} action is executed, or the maximum number of steps, $N=100$, is reached. A successful episode means the agent issues the \emph{stop} action exactly when it reaches the goal (the distance to the target is less than $1m$ and the angle between the current and the target view directions is less than $90^\circ$) within $N$ steps.
Our pipeline is fully automated and does not require human intervention in unknown scenes with new targets.

\section{TARGET-DRIVEN NAVIGATION MODEL}
The goal of the target-driven navigation model is to generate action sequences which are as close as possible to what human would have done in the same situation. In this work, we use a combination of variational generative model and imitation learning to learn a reactive navigation policy, which has shown to outperform some state-of-the-art methods in the context of mapless navigation.
In what follows, we describe how we learn the navigation policy and some additional techniques to facilitate the performance.

\subsection{Navigation Policy}
\label{sec:policy}
Let $\mathcal{S}:\{x_1, a_1,x_2,a_2,\cdots,x_T\}, s.t.~T\leq N$, be the sequence of observations and actions generated by the agent as it navigates to a target $g$. The data is used to learn the target-driven navigation policy $\pi(x_t, g)$, which takes as input a pair of views $(x_t, g)$ and outputs an action $a_t$ required to approach the target observation $g$ from the current observation $x_t$.
We first present $\pi$ by a reactive deep network, which
is trained by minimizing a cross-entropy loss as:
\begin{equation}\label{eq:init}
\mathcal{L}_\pi=E_{a_{t}\thicksim p(a_{t}|x_t, g)}[-\log{\pi(a_t|x_t, g) }]
\end{equation}
where $p$ is the ground-truth action distribution. To minimize the loss, it is common to assume $p$ as a delta function at a ground truth action. However, this assumption is notably violated, since there can be multiple ground truth actions for $(x_t, g)$ and $p$ is inherently multi-modal. When navigation trajectories are longer, more paths may take the agent from the current observation to the target observation, leading to a more difficult multi-modality issue.
Previous works on imitation learning~\cite{watkins2019learning, richter2017safe} typically assume $p$ to be a delta function, which leads to high-variance in gradients during learning and in turn would make learning challenging. Recent deep reinforcement learning models~\cite{zhu2017,wu2018building} require abundant samples to obtain a good empirical estimate of a multi-modal action distribution $p$.
We account for multi-modality by employing a variational generative process. Instead of learning the complex function from visual observations $(x_t, g)$ to action $a_t$ directly, we propose first learning to generate the next observation (NEO) $x_{t+1}$ from $(x_t, g)$ and then learning the mapping from $(x_t, x_{t+1})$ to $a_t$. The mapping is a surjection, which means there is only one appropriate action $a_t$ for $(x_t, x_{t+1})$. In this way, the multi-modality essentially affects the generation of $x_{t+1}$, which is learned by a generative module as~\cite{wuneonav}.

\textbf{Generative Module.} Given the current observation $x_t$, we first model the environment transition dynamics as:

\vspace{-10pt}
\begin{equation}
\label{eq:generative}
p_{\theta}(x_{t+1}, z|x_{t}, a_{t})=p_{\theta}( x_{t+1}|z)p_{\theta}(z| x_{t},a_{t})
\end{equation}

where $p_{\theta}( x_{t+1}, z|x_t, a_t)$ is a parametric model of the joint distribution over the NEO $x_{t+1}$ and a latent variable $z$. To learn the generative model $x_{t+1}\thicksim p_{\theta}(x_{t+1}|x_t, a_t)$, one typically maximizes the marginal log-likelihood $\log p_{\theta}(x_{t+1}|x_t, a_t)$. Since the next action $a_t$ is unknown a \emph{priori} and is inherently determined by the target $g$, we apply variational inference and introduce a distribution $q_{\lambda}(z|x_t, g)$ with parameters $\lambda$ that approximates the true distribution $p_{\theta}(z|x_t, a_t)$. Then we obtain the marginal likelihood of the model:

\vspace{-10pt}
\begin{equation}
\label{eq:lower_bound}
\begin{aligned}
&\log{p_{\theta}(x_{t+1}|x_t,a_t)}=\log{\int_{z}p_{\theta}(x_{t+1},z|x_t,a_t)dz}\\
&=\log{\int_{z}p_{\theta}(x_{t+1},z|x_t,a_t)\frac{q_{\lambda}(z|x_t,g)}{q_{\lambda}(z|x_t,g)}dz}\\
&=\log{E_{z\thicksim q_{\lambda}(z|x_t,g)}[\frac{p_{\theta}(x_{t+1},z|x_t,a_t)}{q_{\lambda}(z|x_t,g)}]}\\
&\geq E_{z\thicksim q_{\lambda}(z|x_t,g)}[\log{\frac{p_{\theta}(x_{t+1},z|x_t,a_t)}{q_{\lambda}(z|x_t,g)}}]
\end{aligned}
\end{equation}

To maximize the marginal likelihood, we maximize its lower bound:

\vspace{-10pt}
\begin{small}
\begin{equation}\label{eq:lower_bound2}
\begin{aligned}
&E_{z\thicksim q_{\lambda}(z|x_t,g)}[\log{\frac{p_{\theta}(x_{t+1},z|x_t,a_t)}{q_{\lambda}(z|x_t,g)}}]\\
&=E_{z\thicksim q_{\lambda}(z|x_t,g)}[\log{\frac{p_{\theta}(x_{t+1}|z)p_{\theta}(z|x_t,a_t)}{q_{\lambda}(z|x_t,g)}}]\\
&=E_{z\thicksim q_{\lambda}(z|x_t,g)}[\log{p_{\theta}(x_{t+1}|z)}+\log{\frac{p_{\theta}(z|x_t,a_t)}{q_{\lambda}(z|x_t,g)}}]\\
&=E_{z\thicksim q_{\lambda}(z|x_t,g)}[\log{p_{\theta}(x_{t+1}|z)}]-\mathcal{K}\mathcal{L}[q_{\lambda}(z|x_t,g)||p_{\theta}(z|a_t,x_t)]
\end{aligned}
\end{equation}
\end{small}
where $\mathcal{K}\mathcal{L}$ denotes the Kullback-Leibler divergence between two distributions, $q_{\lambda}(z|x_t,g)$ and $p_{\theta}(z|a_t,x_t)$.
We design a generative module for maximizing the lower bound, in which
$q_{\lambda}(z|x_t,g)$, $p_{\theta}(z|a_t,x_t)$, and $p_{\theta}(x_{t+1}|z)$ are all parameterized by neural networks.

During training, the navigation tasks to be imitated are provided with a series of ground truth trajectories, e.g., $\{x^{gt}_1, a^{gt}_1,x^{gt}_2, a^{gt}_2,\cdots,x^{gt}_T\}$, which are captured using Dijkstra algorithm.
Therefore, $p_{\theta}(z|a_t,x_t)$ can be estimated as a Gaussian distribution conditioned on the current observation $x_t$ and the ground-truth action $a^{gt}_t$, leading to a \emph{mixture-of-posteriors prior} imposed on the latent distribution for the multi-modality in the generation of the next observation. By minimizing the $\mathcal{K}\mathcal{L}$ divergence, the two distributions, $q_{\lambda}(z|x_t,g)$ and $p_{\theta}(z|a_t,x_t)$, get close to each other, which propels the generation of the next observation, $p_{\theta}(x_{t+1}|z)$,
to be in favour of the navigation task and consistent with the environment transition dynamics meantime. In addition, we empirically approximate $x_{t+1}\thicksim p_{\theta}(x_{t+1}|z)$ using samples $x^{gt}_{t+1}$, that are obtained by the agent after executing $a^{gt}_t$ at $x_t$. From the above, the loss for our generative module is:

\vspace{-10pt}
\begin{equation}\label{eq:gloss}
\mathcal{L}_g=\sum_{t=i}^{T}(\alpha||x^{gt}_{t+1}-x_{t+1}||^2_2+\beta\mathcal{K}\mathcal{L}[q_{\lambda}(z|x_t,g)||p_{\theta}(z|a^{gt}_t,x_t)])
\end{equation}

\textbf{Predictive Control.} Further, to realize robot navigation, we learn a navigation action controller $q_{\varphi}(a_t|x_t,x_{t+1},a_{t-1})$, which predicts the next best action $a_t$ based on the current observation $x_t$, the generated next expected observation $x_{t+1}$ as well as the previous action $a_{t-1}$.
Note that inputting the previous action to our model at each time step could be promising when an agent runs back and forth in a scene.
Given the ground truth action $a^{gt}_t$, the controller is trained by minimizing the standard cross-entropy loss as:

\vspace{-10pt}
\begin{equation}\label{eq:control}
\mathcal{L}_c=E_{a_{t}\thicksim p(a^{gt}_{t}|x_t, g)}[-\log{q_{\varphi}(a_t|x_t, x_{t+1}, a_{t-1})}]
\end{equation}

Integrating the predictive control with the generative module (see Figure~\ref{fig:policy}), the objective of our navigation policy becomes:

\vspace{-10pt}
\begin{equation}
\label{eq:obj_func}
\begin{aligned}
\mathcal{L}_\pi=&\sum_{t=i}^{T}(\alpha||x^{gt}_{t+1}-x_{t+1}||^2_2+\beta \mathcal{K}\mathcal{L}[q_{\lambda}(z|x_t,g)||p_{\theta}(z|a^{gt}_t,x_t)]\\
&+\gamma E_{a_t\thicksim p(a^{gt}_t|x_t, g)}[-\log{q_{\varphi}(a_t|x_t,x_{t+1},a_{t-1})}])
\end{aligned}
\end{equation}
where the hyper-parameter $(\alpha, \beta, \gamma)$ tunes the relative importance of the reconstruction term, the KL divergence term and the predictive control term. The three hyper-parameters are empirically set as $\alpha=0.01$, $\beta=0.0001$, and $\gamma=1$ throughout our experiments.

\subsection{Techniques to Facilitate the Navigation}
\label{sec:tech}
We also investigate three techniques to improve robot navigation performance in the real world. First, we learn the environment dynamics in the feature space as opposed to the raw observation space.
Furthermore, we propose a premature collision prediction module to improve the safety during navigation.
Finally, a target checking module is also designed to issue the \emph{stop} when the agent is near the target.

\begin{figure}[thpb]
\begin{center}
\includegraphics[width=.95\linewidth]{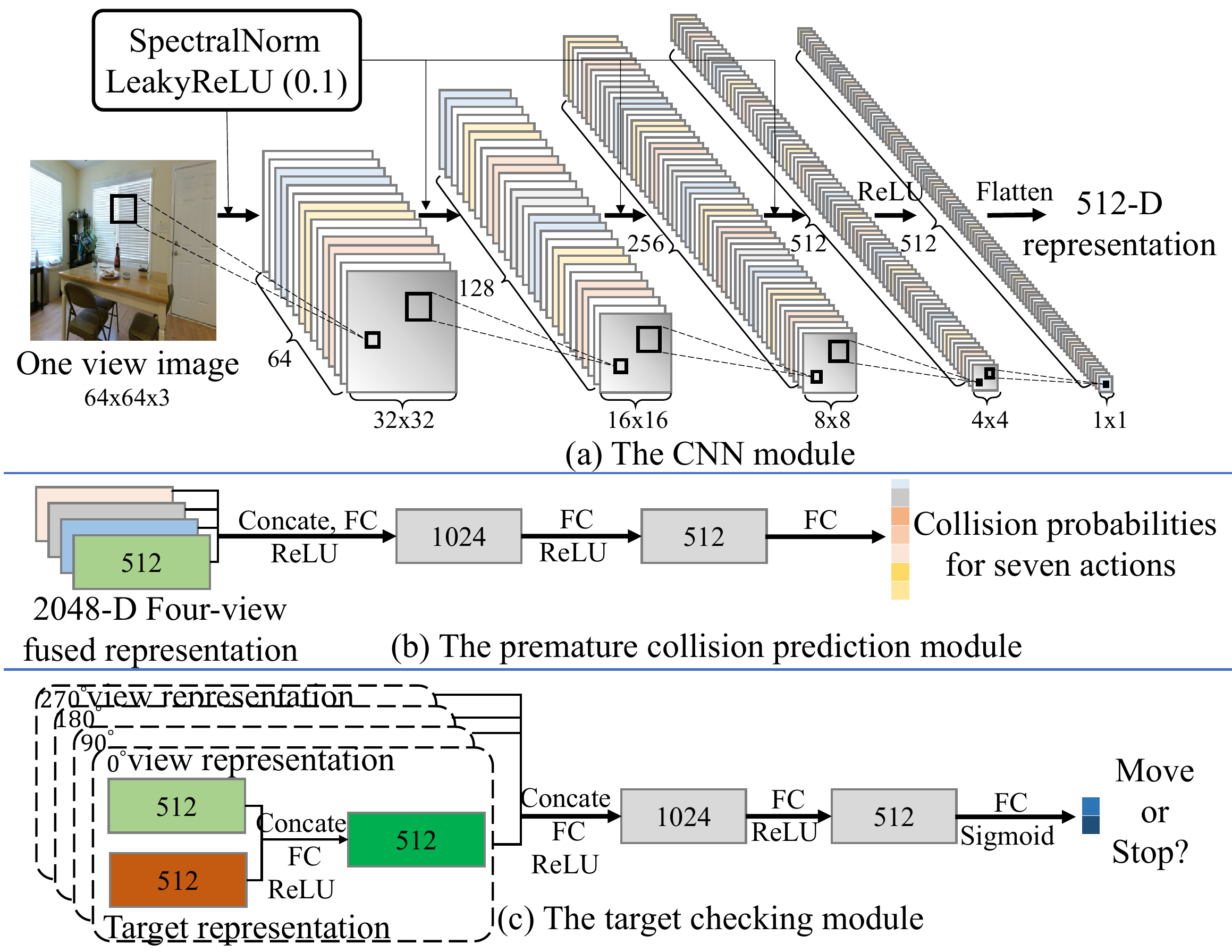}
\end{center}
   \caption{The architectures of our (a) CNN module, (b) premature collision prediction module, (c) target checking module.}
\label{fig:archi}
\end{figure}

\textbf{Feature Space Dynamics.}
\cite{agrawal2016learning} and~\cite{buesing2018learning} have proposed improving the generalization of learning models by learning forward dynamics in the feature space instead of raw observation space.
Following this, we extend our navigation model to make predictions in feature representations of raw observations.
We apply a CNN module $\Phi(\cdot)$ to derive a feature representation from an observation and hence get the current feature $\Phi(x_t)$, the ground truth next state feature $\Phi(x^{gt}_{t+1})$, and the target feature $\Phi(g)$.
We have conducted some experiments on evaluating the choice of the CNN module, e.g., the sophisticated convolutional layer in ResNet$50$~\cite{he2016deep} and VGG$16$~\cite{simonyan2014very} (see Section~\ref{sec:nav_AVD}). Considering both the efficiency and the navigation performance, we design our CNN module in Figure~\ref{fig:archi}(a). The module can compress an RGB or Depth image into a $512$-D feature space. Spectral normalization is used for the first four convolutional layers, which can prevent the escalation of parameter magnitudes and avoid unusual gradients in training~\cite{miyato2018spectral}. The activation function used is LeakyReLU $(0.1)$.

In addition, we directly use the feature after two fully connected (FC) layers of $p(x_{t+1}|z)$, denoted as $f(x_{t+1})$, to help the predictive navigation control $q_{\varphi}$ and update the reconstruction term in Equation~\ref{eq:obj_func}. The final objective with feature space dynamics is as follows:

\vspace{-10pt}
\begin{small}
\begin{equation}
\label{eq:fobj_func}
\begin{aligned}
\mathcal{L}_\pi=&\sum_{t=i}^{T}(\alpha||\Phi(x^{gt}_{t+1})-f(x_{t+1})||^2_2\\
&+\beta \mathcal{K}\mathcal{L}[q_{\lambda}(z|\Phi(x_t),\Phi(g))||p_{\theta}(z|a^{gt}_t,\Phi(x_t))]\\
&+\gamma E_{a_t\thicksim p(a^{gt}_t|x_t, g)}[-\log{q_{\varphi}(a_t|\Phi(x_t),f(x_{t+1}),a_{t-1})}])
\end{aligned}
\end{equation}
\end{small}
\textbf{Premature Collision Prediction.}
We propose incorporating an auxiliary module into our navigation policy in order to
promote more robust learning, and ultimately safer navigation performance for our agent. This auxiliary module is a multilayer perceptron (MLP) downstream of the CNN module of our navigation policy, which provides the collision probability of all actions in $\mathcal{A}$ over the current four-view observation $x_t$ (see Figure~\ref{fig:archi}(b)).
We refer to this as the premature collision prediction module, $ca_t\thicksim \textup{CPM}(ca_t|x_t)$, leading to a multi-label classification loss term, which is specified as follows:
\begin{equation}\label{eq:target}
\mathcal{L}_{cp}=-\sum_{t=i}^{T}\sum_{ca_t\in\mathcal{A}}p(ca^{gt}_t|x_t)\log{\textup{CPM}(ca_t|x_t)}
\end{equation}
where $p(ca^{gt}_t|x_t)$ is a delta function at $ca^{gt}_t$, which is provided by the interaction between the agent and the environment.

Given the state representation of the current observation, $\Phi(x_t)$, the auxiliary module can be summarized as predicting the collision probability of all possible actions at this time step.
The module shares the same CNN module as the navigation policy network.
We believe this forces the CNN module to learn low-level representations that are useful for both the navigation task and collision avoidance.

\textbf{Target Checking.}
The target checking module is especially critical for robot navigation in the real world, which enables a robot to figure out if the current target is reached. This process is simple given knowledge of the true physical state, but difficult when working with visual observations. Aside from the usual challenges of visual recognition, the significant training data imbalance further complicates the target checking task~\cite{zhao2018triangle}, since we only have one positive example of \emph{stop} action at the end of each trajectory, while all the other steps are negative examples for \emph{not stop}.

We pose the target checking as a binary classification problem and design a target checking module $\textup{TCM}(\cdot)$ that takes in the current four-view observation concatenated with the target image and predicts whether the agent reaches the target position, denoted as $sa_t\thicksim \textup{TCM}(sa_t|x_t, g)$. The target checking module is jointly trained with our navigation policy. It shares the CNN and the feature fusion parts with our navigation policy, which provides a $2048$-D fused feature vector at each time step.
The vector finally passes through a MLP to output the probability of the target being reached.
A detailed topology of the module is pictured in Figure~\ref{fig:archi}(c).
Similar to~\cite{wuneonav}, for alleviating the effect of data imbalance, we guarantee that approximately $\frac{1}{|\mathcal{A}|}$ training tasks are near navigation targets, of which the optimal action is \emph{stop} at the start location.
We train the target checking module using a binary cross-entropy loss defined as:
\begin{equation}\label{eq:target2}
\mathcal{L}_{tc}=E_{sa_t\thicksim p(sa^{gt}_t|x_t,g)}[-\log{\textup{TCM}(sa_t|x_t,g)}]
\end{equation}
where $p(sa^{gt}_t|x_t,g)$ is a delta function at $sa^{gt}_t$, which is supervised by the environment.

Subsequently, we introduce $\zeta$ and $\eta$ to integrate both the premature collision prediction module and the target checking module into our navigation policy learning. The two weights control the strength of the two auxiliary loss terms. We experiment with several different constants and $\zeta=0.4,~\eta=0.4$ are finally determined.
The goal of automatically computing the optimal weights for an arbitrary environment is a good topic for future work.
Our overall navigation objective is given:
\begin{equation}\label{eq:all}
\mathcal{L}=\mathcal{L}_{\pi}+\zeta\mathcal{L}_{cp}+\eta\mathcal{L}_{tc}
\end{equation}

At test time, our model outputs a navigation command given the current observation, the target view and the previous action progressively.
This will drive the robot toward the eventual target while avoiding some static obstacles in unseen scenes, without relying on any maps or location services.




\section{Experiments and Discussions}
\label{sec:result}
We evaluate our model by testing on both synthetic and real-world $3$D navigation tasks. We present our navigation performances in the context of the different choices we made in our design, as well as comparing with some alternate methods.
The key characteristic of a good navigation policy is that it generalizes to unseen scenes and new targets while remaining robust to irrelevant parts of visual observations.

\subsection{Experimental Setup}
\label{subsection:setup}
\textbf{Alternatives.}
We compare the navigation performance to the following alternative models:
(1) \textbf{NeoNav} is our previous work~\cite{wuneonav}.
(2) \textbf{TD-A3C} is  the abbreviation of the baseline~\cite{zhu2017}. To evaluate the generalization to different targets in unknown scenes, we just keep one scene specific layer of the network and train on all training scenes.
(3) \textbf{TD-A3C-IL} incorporates IL into TD-A3C for the sample-inefficiency in RL. Furthermore, for a fair comparison, the variant is facilitated with $\textup{CPM}$ and $\textup{TCM}$, and uses the same CNN module and input as ours.
(4) \textbf{G-LSTM-A3C-IL} is a variant of TD-A3C-IL using more advanced architectures with LSTM from~\cite{wu2018building}.
(5) \textbf{GSP} is a goal-conditioned skill policy in~\cite{pathak2018zero}, which learns an inverse dynamics model based on some demonstrated way-points from an expert and predicts the next state feature as an auxiliary task for control. We reimplement the work based on their provided code\footnote{https://github.com/pathak22/zeroshot-imitation} and train it in our setup for a fair comparison.
(6) \textbf{LSTM-A3C-KG-A}~\cite{lv2020improving} uses knowledge graph and attention mechanism both to form spatial reasoning and guide policy search.

\textbf{Implementation Details.}
Our model is trained and tested on a PC with $12$ Intel(R) Xeon(R) W-$2133$ CPU, $3.60$ GHz and a Geforce GTX $1080$ Ti GPU.
We use an RMSprop optimizer~\cite{tieleman2014rmsprop} to update our model with a learning rate of $1e^{-4}$ and a smoothing constant of $0.99$.
During training, each update is based on $60$ time steps from $6$ random trajectories,
each of which is generated by randomly selecting a scene, a start and a target from our training split.

\textbf{Evaluation Metrics.}
When sampling evaluation tasks,
we consider the ratio of the shortest path distance to the Euclidean distance between the start and goal positions of a task, proposed by~\cite{savva2019habitat} to benchmark navigation task difficulty. In each evaluation, we compute the percentage $P$ (lower is more challenging) of the tasks that have a ratio within the range of $[1, 1.1]$.
In addition, we adopt two main evaluation metrics in our experiments: success rate (SR) and success weighted by path length (SPL)~\cite{yang2018visual}. For each of the $K$ navigation evaluation tasks, let $S_i$ be a binary indicator for successful navigation or unsuccessful navigation.
$l_i$ and $p_i$ denote the length of the shortest path and the actual executed path of the $i$-th task, respectively.
Success rate is the fraction of tasks in which the agent reaches the target successfully within limited time steps: $\emph{SR}=\frac{1}{K}\sum^{K}_{i=1}S_i$.
SPL considers both the success indicator and the length of the executed path:
$\emph{SPL}=\frac{1}{K}\sum^{K}_{i=1}S_i\frac{l_i}{\max(l_i,p_i)}$.
The higher this value, the faster, on average, the agent approaches the target.




\subsection{$3$D Navigation in AVD}
\label{sec:nav_AVD}
We first conduct our experiments based on the training and testing splits of AVD~\cite{mousavian2019visual}.
The input visual resolution is $64\ast64$.
For each training scene, we choose fifteen different views as the targets by default, each of which contains a common object, such as a dining table, a sofa, a television, etc. The start position of a navigation agent can be randomly sampled across the scene.
The training times of our model, NeoNav, TD-A3C-IL, G-LSTM-A3C-IL, and GSP are all about $24$ hours for RGB or depth input.
TD-A3C requires double the training time.
For evaluation, two kinds of settings are considered here, $\{\emph{Seen}~\emph{environments},~\emph{Novel}~\emph{targets}\}$ and $\{\emph{Unseen}~\emph{environments},~\emph{Novel}~\emph{targets}\}$. Each evaluation contains $1000$ different navigation tasks ($P=15.0\%$).
The target views of these evaluation tasks, which are different from the training target views, are randomly sampled.

\textbf{Ablations.}
We first evaluate how the performance is affected by changing the input modality of our model.
We train our model with RGB inputs which leads to the SR/SPL (in $\%$) of $21.3/6.9$ and the model with RGBD inputs leads to the SR/SPL of  $26.6/8.4$ on the evaluation from unseen environments.
Compared to RGB, depth images contain rich geometry information which benefits a powerful reasoning about the surrounding layout, leading to better navigation policies ($28.7/8.8$).
However, learning the effective combination of features from RGB and depth images may further complicate the navigation decision making, resulting in worse performance.
Hence, unless explicitly stated otherwise, we use depth by default.


An ablation study on the CNN backbone is provided based on the tasks from unseen environments.
The navigation performance (SR and SPL in \%) of our re-implementations of VGG$16$ and ResNet$50$ are: $25.9/8.5$ and $30.2/10.9$, respectively.
These are similar to \textbf{Ours} ($28.7/8.8$) with the CNN design in Section~\ref{sec:tech}.
However, the training time of VGG$16$ is at least twice that of our algorithm and the training of ResNet$50$ is three times longer than training our current architecture.
Hence, we suggest our design considering both the navigation performance and the training time.

Our navigation policy exploits a combination of a variational generative model and imitation learning and is augmented with two auxiliary modules, including premature collision prediction, and target checking.
We systematically ablate the components to quantitatively review the importance:
(1) \textbf{Ours-NoVG} removes the variational generative module and predicts navigation actions directly based on the current observation and the target.
(2) \textbf{Ours-NoCP} predicts navigation actions without prematurely considering a collision at each step.
(3) \textbf{Ours-NoTC} learns to output a stop action by the policy rather than a target checking module.

\begin{figure}[thpb]
\begin{center}
\includegraphics[width=\linewidth]{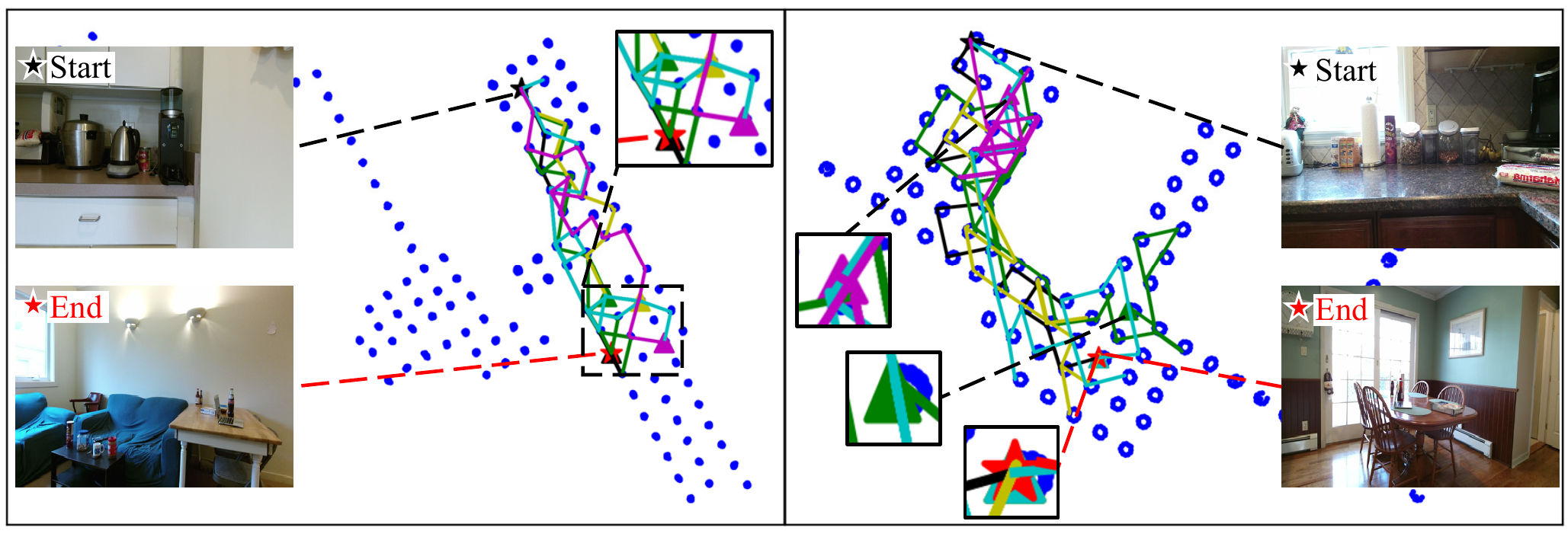}
\end{center}
   \caption{Visual comparison of navigation paths. Blue dots represent the reachable positions in the scenes. Black stars and red stars denote starting and goal points, respectively. Ours, NeoNav, Ours-NoVG, Ours-NoCP, and Ours-NoTC choose the black, the yellow, the magenta, the cyan and the green paths, respectively.
   Triangles in different colors represent end points of different models. Some triangles can be overlapped with others (e.g., the black and the yellow triangles both overlap with the cyan triangle in the second task).
   Only our agent successfully navigates to both the targets.
 }
\label{fig:apath}
\end{figure}

As shown in Table~\ref{tab:ablation1}, our navigation pipeline shows $16.3\%$\footnote{$\textup{Value}=\frac{(C_{\textup{Ours-Pre}}-C_{\textup{Ours}})}{C_{\textup{Ours-Pre}}}$, where $C$ is for the collisions in Table~\ref{tab:ablation1}} reduction in average collisions, $11.2\%$ improvement in average SR, and $4.4\%$ improvement in average SPL over our prior navigation algorithm (NeoNav)~\cite{wuneonav}, which demonstrates the effectiveness of our novel designs in unseen scenes.
Ours-NoVG ignores the multi-modality during navigation by directly learning the complex connection from visual observations to actions, which
is difficult to generalize to new navigation tasks.
Ours-NoCP demonstrates worse static collision avoidance during navigation.
The comparison between Ours-NoTC and Ours shows that the training data imbalance problem significantly affects robot navigation learning, which can be alleviated by additional target checking.
We also visualize navigation trajectories of these models in Figure~\ref{fig:apath}.
Ablation models fail to reach both the targets.
In contrast, our proposed agent performs best in terms of both path quality and navigation success.


\begin{table}[h]
\centering
\caption{Ablation study on model structure based on the average navigation performance (SR and SPL in \%, average collisions for a task) on AVD with depth input.
\label{tab:ablation1}}
{\begin{tabular}{c|c|c|c|c}
\cline{1-5}
\hline
Environment  &Model  &SR &SPL &Collisions\\
\hline
 &NeoNav~\cite{wuneonav} &  17.5 & 4.4 & 57.5\\\cline{2-5}
Unseen&Ours-NoVG&  22.6 & 8.4 & 50.1\\\cline{2-5}
Environments&Ours-NoCP&  27.1 & 8.4& 56.2\\\cline{2-5}
P=15.0\%&Ours-NoTC& 19.8 & 6.3 & 51.3\\\cline{2-5}
&Ours&  \textbf{28.7} & \textbf{8.8}& \textbf{48.1}\\\cline{2-5}
\hline
\end{tabular}}
\end{table}

\begin{table}[h]
\centering
\caption{Navigation performance (SR and SPL in \%) for different number of training target views from AVD with depth input.
\label{tab:table1}}
\scalebox{0.90}{\begin{tabular}{c|c|c|c|c}
\cline{1-5}
\hline
Environment& Seen &\multicolumn{3}{ |c }{Unseen}\\\cline{1-5}
\hline
Target Views $\sharp$  & 120 & 120& 240 & 360 \\\cline{1-5}
\hline
Random& 1.4 /  0.8 & 2.8 / 1.8 & 2.8 / 1.8 & 2.8 / 1.8\\\cline{1-5}
TD-A3C~\cite{zhu2017}&  26.3 /  7.2 &6.4 / 3.4 & 7.9 /  4.1 & 8.1 /  4.0\\\cline{1-5}
TD-A3C-IL~\cite{zhu2017}& 47.2 /  20.1 &20.7 / 6.7 & 23.4 /  7.1 & 26.5 /  8.6\\\cline{1-5}
G-LSTM-A3C-IL~\cite{wu2018building}&39.6 /  11.2& 20.3 / 6.1 & 22.3 / 6.3 & 27.2 / 7.4 \\\cline{1-5}
GSP~\cite{pathak2018zero}& \textbf{52.3} /  \textbf{25.3}&  23.4 / 6.7 & 25.1 /  9.7 & 31.8 /  10.7 \\\cline{1-5}
Ours& 49.3 / 23.6 &  \textbf{28.7} / \textbf{8.8} & \textbf{31.6} / \textbf{10.3} & \textbf{33.3} / \textbf{12.0} \\\cline{1-5}
\hline
\end{tabular}}
\end{table}

\textbf{Comparisons.}
Table~\ref{tab:table1} summarizes the results of our proposed model and the alternatives.
All learning models get better performances when tested on seen scenes than on unseen scenes.
The performance degrades drastically for both the baselines and our proposed models in unseen scenes.
This indicates that all models do not have a deep understanding of navigation tasks and environments.
We assume this is because the accessible scenes are limited and highly discretized during training, impeding the understanding of the real indoor environments characterized by high complexity and continuity. We also evaluate the navigation performance of these models, when trained with RGB inputs (see Table~\ref{tab:rgb}).
We find that depth information consistently improves the navigation performance for all models in unseen environments.

\begin{table}[h]
\centering
\caption{Average navigation performance (SR and SPL in \%) comparisons on AVD with RGB inputs.
\label{tab:rgb}}
\scalebox{0.90}{\begin{tabular}{c|c|c|c|c}
\cline{1-5}
\hline
  &\multicolumn{2}{ |c }{Seen}&\multicolumn{2}{ |c }{Unseen} \\\cline{2-5}
Model  &SR &SPL &SR &SPL\\
\hline
Random &  1.4 & 0.8 & 2.8  &  1.8 	\\\cline{1-5}
TD-A3C&  20.7 & 5.1 & 5.1 &  2.9 \\\cline{1-5}
TD-A3C-IL&  45.2 & 19.7 & 18.2 &  5.9 \\\cline{1-5}
G-LSTM-A3C-IL&  31.7 & 9.5 & 17.9 &  5.5 \\\cline{1-5}
GSP&  47.9 & 14.3 & 19.3 &  5.5 \\\cline{1-5}
Ours&\textbf{54.6} & \textbf{23.5} & \textbf{21.3}&  \textbf{6.9} \\\cline{1-5}
\hline
\end{tabular}}
\end{table}

\begin{figure}[thpb]
\begin{center}
\includegraphics[width=\linewidth]{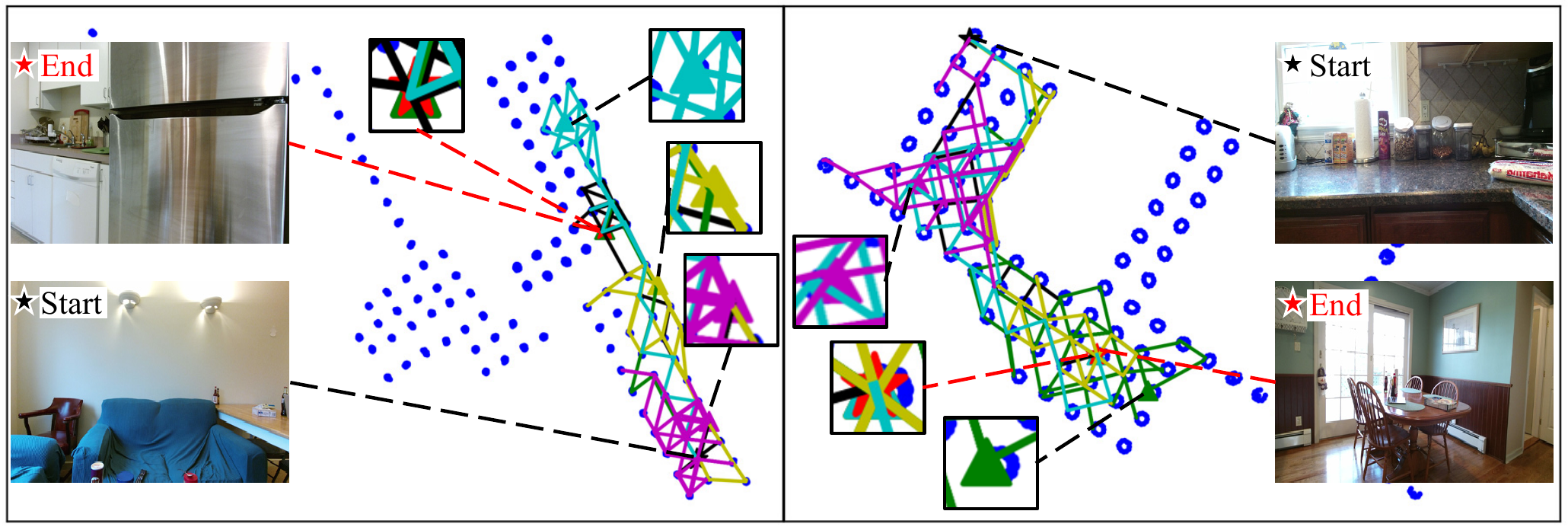}
\end{center}
   \caption{Visual comparison of navigation paths. Blue dots represent the reachable positions in the scenes. Black stars and red stars denote starting and goal points, respectively. Triangles in different colors represent end points of different models.  TD-A3C, TD-A3C-IL, G-LSTM-A3C-IL, and GSP choose the magenta, the green, the cyan and the yellow paths, respectively. Our agent takes the black paths and is able to successfully navigate to the goals in the two cases.}
\label{fig:pathcomp}
\end{figure}

In addition, our proposed navigation pipeline generally outperforms these methods in terms of both path quality and success rate.
TD-A3C is originally designed for scene-specific policy learning and thus lacks the generalization ability to unseen scenes.
Moreover, dealing with sparse rewards is challenging in RL.
TD-A3C-IL that combines IL, the proposed $\textup{CPM}$ and $\textup{TCM}$ together achieves significantly
higher performance than the pure A3C method.
This indicates that imitation learning and some proposed designs have a significant impact on accelerating the learning rates of navigation agents.
The model, G-LSTM-A3C-IL, a direct application of LSTM on TD-A3C-IL, does not yield
sensible performance due to the limited training data.
Moreover, TD-A3C-IL and G-LSTM-A3C-IL both learn the complex function directly from visual observations to navigation actions, which is tough.
This is due to the multi-modality of navigation actions, leading to the weak correlations between visual observations and actions.
GSP addresses the multi-modality with their novel forward consistency,
which makes the GSP-predicted action consistent with the ground-truth action both leading to the next state that benefits a navigation task.
In contrast, our method learns to imagine the next observation from the current observation and the target, and then learns the mapping from the difference between the imagined and the current observations to the navigation action.
This transfers the multi-modality to the generation of NEO, disposed by a variational generative module, and keeps the navigation action prediction process a surjection, which guarantees the strong correlation.
Figure~\ref{fig:pathcomp} shows the agent trajectories by these models for two navigation tasks from unseen scenes. Only our model successfully navigates the agent to the targets in these two cases.

We also evaluate the navigation performance improvements of these models, when trained on increasing numbers of target views from the training split in Table~\ref{tab:table1}. The evaluation is based on the $1000$ navigation tasks from the unseen environments. As can be seen, all the models show increasing SRs and SPLs with increasing numbers of training target views. GSP presents a faster rate of growth than others, indicating that having more targets is advantageous for improving the learning capability of agents.
Furthermore, our model invariably achieves the best results which indicates the data efficiency of the whole architecture.

\begin{figure}[thpb]
\begin{center}
\includegraphics[width=.85\linewidth]{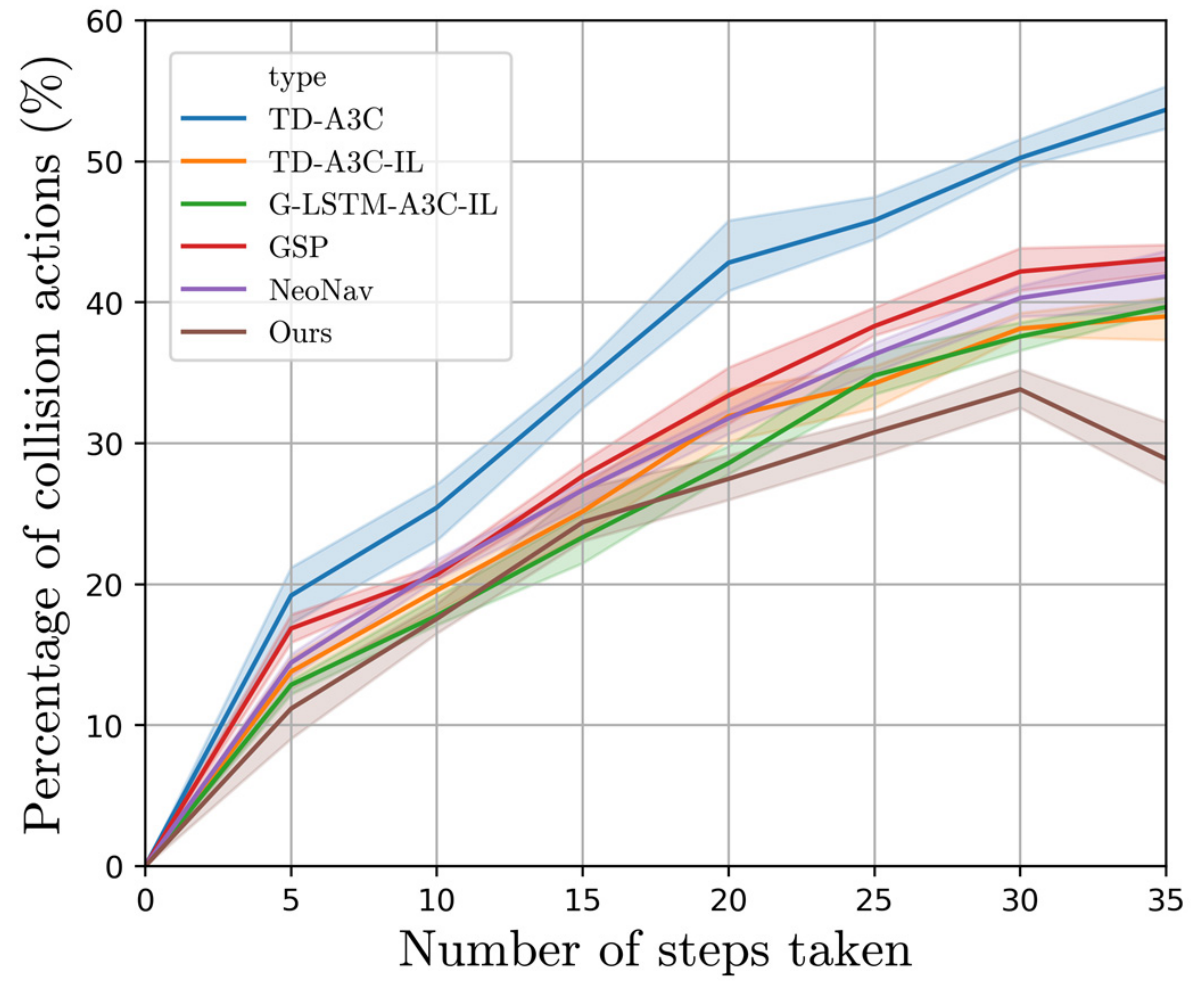}
\end{center}
   \caption{The collision action percentages of learning models as the navigation proceeds. We report the average values over five runs with standard deviations shown in error bands. The most notable observation is that the collision action percentage of our method decreases at time step $35$, indicating that fewer collisions occur during the time interval (30, 35].}
\label{fig:colli}
\end{figure}

In all the experiments, when a static collision occurs, a navigation agent will make a new action decision which may guide the agent out of the dilemma or have it stay put until running out of time, e.g., $100$ steps. We evaluate the collision avoidance capability of these models by computing the ratio of collisions as the navigation proceeds based on the $1000$ navigation tasks from unseen environments with depth inputs. As shown in Figure~\ref{fig:colli}, TD-A3C has the worst performance. This maybe due to the entropy regularisation penalty during training~\cite{jaderberg2016}, which improves the exploration ability of the model leading to less attention on the obstacles in the environments.
TD-A3C-IL, G-LSTM-A3C-IL, and Ours present better static collision avoidance performances than TD-A3C, GSP and Ours-Pre.
We owe it to the premature collision prediction module, since the three models are all facilitated with the module, which encourages the agent to learn to sense static obstacles before acting.

\begin{figure*}
\begin{center}
\includegraphics[width=.9\linewidth]{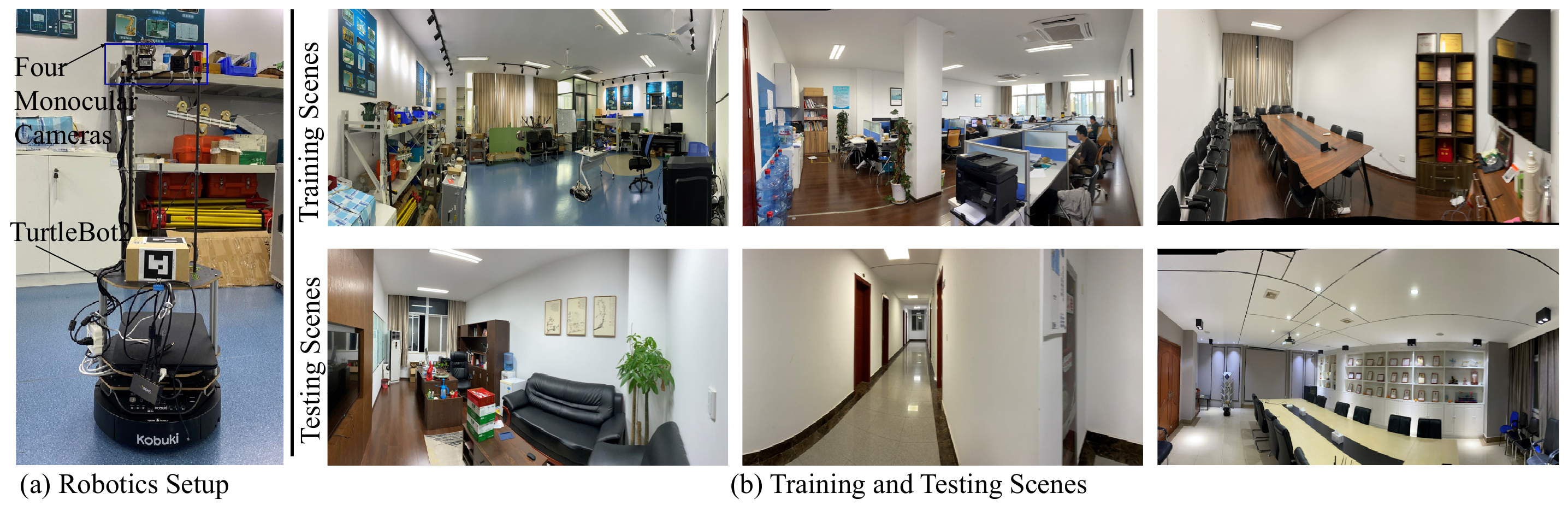}\vspace{-12pt}
\end{center}
   \caption{The robotics system setup and the real-world scenes for training and testing.
 }
\label{fig:robot}\vspace{-12pt}
\end{figure*}

\subsection{$3$D Navigation in AI2-THOR}
We further adapt our method to compare it with LSTM-A3C-KG-A~\cite{lv2020improving},
which integrates a $3$D knowledge graph and sub-targets into a classic deep reinforcement learning framework to boost navigation performance.
The experiment is conducted on all the kitchen rooms of AI2-THOR with the same training/testing split and success criterion as~\cite{lv2020improving}.
We randomly choose navigation tasks from the training/testing split,
and all the initial locations are at least $10$ steps away from the targets.
The performances of two methods are presented in Table~\ref{tab:aicom1}.
The results indicate that both methods are prone to over-fitting.
However, our method shows $6.23\%$ and $14.89\%$ improvements in average SR and SPL compared to LSTM-A3C-KG-A, when evaluated on unseen scenes.
We suggest that our variational generative module facilitates generalizable learning for navigation, which can infer useful information from the perceptible environment.

\begin{table}[h]
\centering
\caption{
Average navigation performance (SR and SPL in \%) comparison on AI2-THOR with RGB input.
\label{tab:aicom1}}
\scalebox{0.90}{\begin{tabular}{c|c|c|c}
\cline{1-4}
\hline
  &Seen scenes,  &Seen scenes, &Unseen scenes \\
  &Seen targets  &Unseen targets & \\
\hline
LSTM-A3C-KG-A~\cite{lv2020improving} &  \textbf{98.44} / 52.58& 44.25 / 14.89 & 41.09 / 7.20\\\cline{1-4}
Ours&  91.58 / \textbf{75.16} & \textbf{86.33 / 67.04} & \textbf{47.32 / 22.09}\\\cline{1-4}
\hline
\end{tabular}}
\end{table}

\subsection{$3$D Navigation in Real-World Indoor Scenes}
We have evaluated our approach on a real-world dataset thus far.
To validate the generalization to real-world settings, we employ a mobile robot, TurtleBot,
equipped with four onboard monocular cameras for sensing RGB images.
In the TurtleBot settings, the action space is also set to be consistent with $\mathcal{A}$ by using of velocity control. The move action is approximately $0.5m$ translation and the rotate action is approximately $30$ degrees of rotation.
The \emph{move right/left} action is complex due to the movement direction restrictions of TurtleBot. For example, we use a series of combinatorial motions, $\{\emph{Rotate~right}~90^\circ,\emph{Move~forward}~0.5m,\emph{Rotate~left}~90^\circ\}$ to produce the \emph{move right} action in $\mathcal{A}$.
We first use our robot to collect data from three training scenes in an academic building in the same way as~\cite{mousavian2019visual} (see Figure~\ref{fig:robot}, \emph{the dataset will be made publicly available}), and then transfer our navigation model from AVD to the three scenes.
The motivation for this setup is to help the agent become familiar with the general layouts of office or laboratory environments, and weaken the effect of robot type as well.
We also test the robot in three more never-before-encountered office scenes for both the cross-target and cross-scene generalization evaluation.

\begin{figure*}[thpb]
\begin{center}
\includegraphics[width=.8\linewidth]{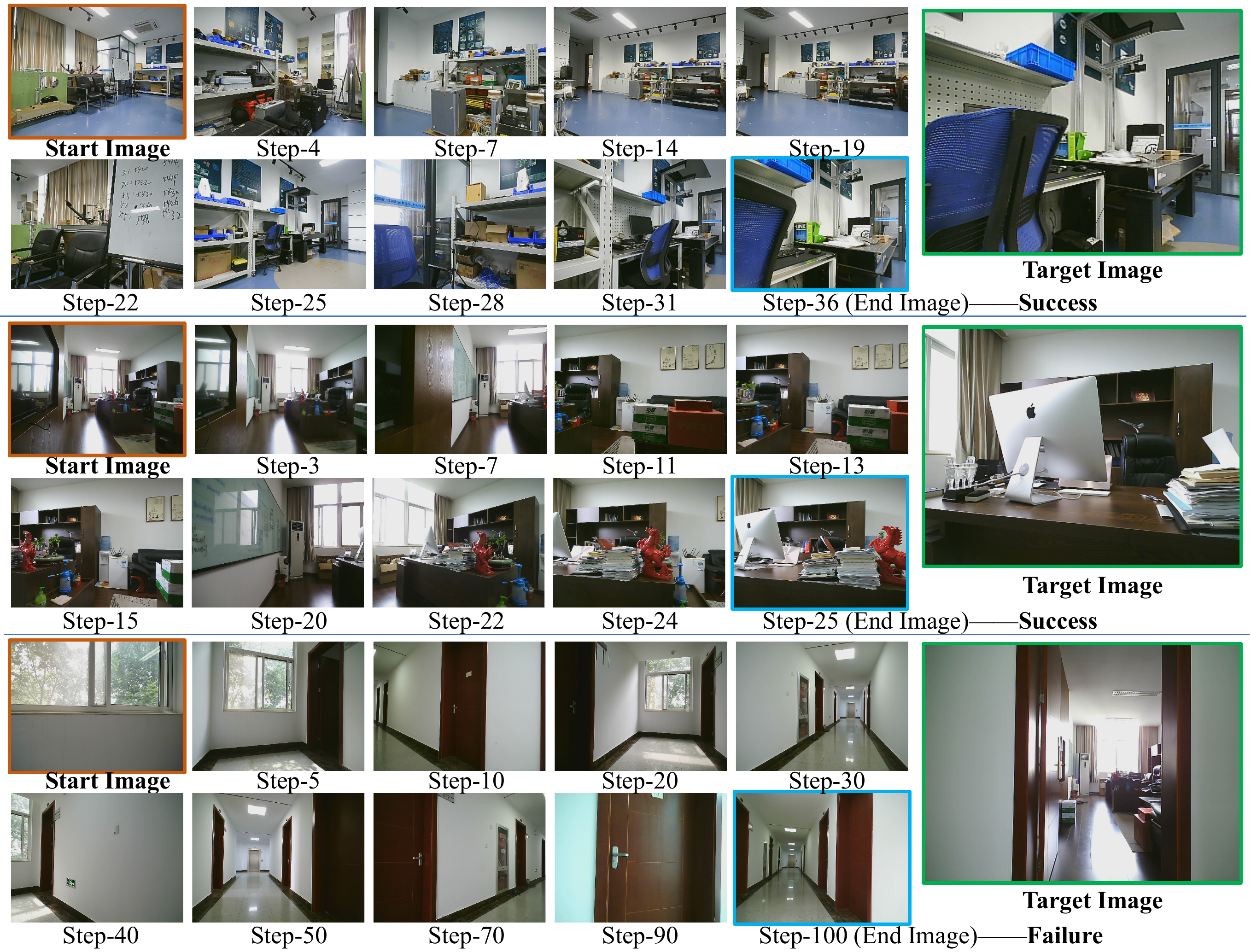}\vspace{-12pt}
\end{center}
   \caption{Visualization of the three trajectories of a TurtleBot to reach targets (in green) from the start images (in orange). The robot manages to reach the locations (in cyan) near the targets in the first two navigation tasks and fails in the scene with many repetitions (e.g., doors in a corridor).}
\label{fig:tra}\vspace{-12pt}
\end{figure*}

We first evaluate the cross-target generalizations of navigation models in the same training scenes.
We start the TurtleBot from different starting locations and orientations and set up $50$ navigation tasks.
In addition, we test the robot in another three different office scenes that it has never encountered before for both the cross-target and cross-scene generalization evaluation by randomly setting up another $50$ navigation tasks.
We judge the navigation to be successful if the robot stops near the target, and consider it a failure if the robot collides with an obstacle or does not reach the goal within $100$ steps.
The performance is measured by the success rate over all navigation tasks.
We also choose four tasks from the evaluation and show the results when using different models.
As shown in Table~\ref{tab:re1},
TD-A3C-IL and G-LSTM-A3C-IL both struggle to generalize to unseen scenes using RGB,
due to the high complexity of real indoor environments.
The performances on both training scenes and testing scenes of GSP and our model are similar to the evaluation results on AVD.
In general, our model outperforms these methods in generalizing to new targets or new scenes with novel layouts.
\begin{table}[h]
\centering
\caption{Average navigation performance comparisons in real world with RGB input (SR in \%).
\label{tab:re1}}
\scalebox{0.80}{\begin{tabular}{c|c|c||c|c|c||c}
\cline{1-7}
\hline
  &\multicolumn{3}{ |c }{Three training scenes}&\multicolumn{3}{ |c }{Three testing scenes} \\\cline{1-7}
Model   &Task-1&Task-2 &SR  &Task-1 &Task-2 &SR\\
\hline
TD-A3C-IL~\cite{zhu2017}&  Collide & 52~Steps  & 44.0   &  Collide & Fail &  4.0 \\\cline{1-7}
G-LSTM-A3C-IL~\cite{wu2018building}&  Fail&  72~Steps & 40.0   &  Fail&  Fail &  6.0\\\cline{1-7}
GSP~\cite{pathak2018zero}  &  65~Steps&  41~Steps & 58.0   &  Collide &  Fail & 22.0 \\\cline{1-7}
Ours & 51~Steps&  38~Steps &62.0   &  Fail&  51~Steps & 30.0 \\\cline{1-7}
\hline
\end{tabular}}
\end{table}

In addition, we observe that our model achieves great performance when the target view contains some distinct objects, which happen to be in the front view of the start location.
However, there is a high probability that the agent gets stuck in the corner and thrashes around in space without making progress when the initial view of the agent and the target image have no overlap.
See Figure~\ref{fig:tra} for three front-view trajectories of TurtleBot generated by our method.
Moreover, the robot exhibits significantly oscillatory and jerky motions during navigation since our method and all alternatives only predict discrete action commands.
The testing time of our model for each navigation decision is about $0.03s$.
However, the time for the robot to finish each locomotion is much longer.
For example, the \emph{move right} action in $\mathcal{A}$ is converted to rotate right at $45^\circ/s$ for $2s$, move forward at $0.25m/s$ for $2s$, and rotate left at $45^\circ/s$ for $2s$. The saltatorial velocity control results in jerky motions.
Extension to continuous velocity control would make the method applicable in realistic environments.
\emph{Videos are available in the supplementary material.}

\section{Conclusions and future work}
\label{sec:con}
In this work, we present a navigation pipeline for target-driven visual navigation, which does not rely on any maps or localization services at runtime and is purely based on the visual input and a target image.
In contrast to most learning-based navigation methods, we design a generative module before predictive navigation control.
The key idea is to transfer the multi-modality in visual navigation control to the intermediate generative process, which is dealt with in a variational model.
This transfer strengthens the connection between visual observations and navigation actions, thus improving the learning capability of our agent and leading to better generalization to new targets or scenes.
In addition, we investigate three techniques to facilitate navigation, which further improves both the cross-scene and cross-target generalization of our agent in the real world.

One thing to note is that, in the current framework, we do not place attention on relevant areas of the visual input, which can allow the system to more rapidly
detect useful information for decision making.
Our future work will explore perceptual control in feature space during navigation learning, and will evaluate it in more complex environments.

\bibliographystyle{IEEEtran}

\bibliography{bibtex}

\end{document}